\documentclass{article}
\usepackage{ijcai22}

\usepackage{times}

\usepackage[dvipsnames]{xcolor}
\usepackage{soul}
\usepackage{url}

\usepackage[hidelinks,colorlinks,citecolor=MidnightBlue]{hyperref}
\usepackage[utf8]{inputenc}
\usepackage[small]{caption}
\usepackage{graphicx}
\usepackage{amsmath}
\usepackage{booktabs}
\usepackage{graphicx}
\usepackage{amssymb}
\usepackage{booktabs}

\usepackage{makecell}
\usepackage{blindtext}
\usepackage{bbm}
\usepackage{enumitem}
\usepackage{epsfig}
\usepackage[ruled]{algorithm2e}
\usepackage{algorithmic}
\usepackage{multirow}
\usepackage{color}
\usepackage{xcolor}
\usepackage{color, colortbl}
\usepackage{array}
\usepackage[english]{babel} 
\usepackage{amsmath,amsfonts,amsthm} 
\usepackage{mathtools}
\usepackage{pifont}
\usepackage[capitalize]{cleveref}

\newcommand{\ie}{\textit{i.e.}}
\newcommand{\eg}{\textit{e.g.}}
\newcommand{\wrt}{\textit{w.r.t.}}
\newcommand{\argsort}{\operatorname{argsort}}
\newcommand{\softsort}{\operatorname{softsort}}
\newcommand{\softmax}{\operatorname{softmax}}
\newcommand{\sort}{\operatorname{sort}}

\newcommand{\s}{\mathbf{s}}
\newcommand{\ds}{\operatorname{d}_\text{s}}
\newcommand{\f}{f_\theta}
\newcommand{\xx}{\mathbf{x}}

\newcommand{\bb}{\mathbf{b}}
\newcommand{\zz}{\mathbf{z}}
\newcommand{\hh}{\mathbf{h}}
\newcommand{\sss}{\mathbf{s}}
\newcommand{\func}[1]{\textcolor{RoyalBlue}{\mathtt{#1}}}
\newcommand{\comment}[1]{\textcolor{gray}{\#~\mathtt{#1}}}
\newcommand{\variable}[1]{\textcolor{black}{\mathtt{#1}}}
\newcommand{\param}[1]{\textcolor{BrickRed}{\mathtt{#1}}}

\newcommand{\XX}{\mathbf{X}}
\newcommand{\ZZ}{\mathbf{Z}}

\newcommand{\BB}{\mathbf{B}}
\newcommand{\SSS}{\mathbf{S}}
\newcommand{\EE}{\mathbf{E}}
\newcommand{\ee}{\mathbf{e}}

\newcommand{\ZZa}{\tilde{\mathbf{Z}}}
\newcommand{\HHa}{\tilde{\mathbf{H}}}
\newcommand{\BBa}{\tilde{\mathbf{B}}}

\newcommand{\HHb}{\hat{\mathbf{H}}}
\newcommand{\BBb}{\hat{\mathbf{B}}}

\newcommand{\pp}{\mathbf{p}}
\newcommand{\PP}{\mathbf{P}}

\newcommand{\highlight}[1]{\colorbox{red!20}{$\displaystyle #1$}}

\newcommand{\chighlight}[2]{\colorbox{#1}{$\displaystyle #2$}}
\newcommand{\cbrace}[3]{\color{#1}\underbrace{\color{black}#2}_{#3}}

\title{\textit{Learning to Hash} Naturally Sorts}

\author{

Jiaguo Yu$^1$$^*$\and
Yuming Shen$^2$$^*$\and
Menghan Wang$^3$\and
Haofeng Zhang$^{1}$$^\dagger$\and
Philip H.S. Torr$^2$
\\
\affiliations
$^1$Nanjing University of Science and Technology\\
$^2$University of Oxford\\
$^3$eBay\\
}

\begin{document}

\maketitle
\def\thefootnote{*}\footnotetext{Equal contribution.}\def\thefootnote{\arabic{footnote}}
\def\thefootnote{$\dagger$}\footnotetext{Corresponding author.}\def\thefootnote{\arabic{footnote}}
\begin{abstract}
\textit{Learning to hash} pictures a list-wise sorting problem. Its testing metrics, \eg, mean-average precision, count on a sorted candidate list ordered by pair-wise code similarity. 
However, scarcely does one train a deep hashing model with the sorted results end-to-end because of the non-differentiable nature of the sorting operation. This inconsistency in the objectives of training and test may lead to sub-optimal performance since the training loss often fails to reflect the actual retrieval metric. In this paper, we tackle this problem by introducing Naturally-Sorted Hashing (NSH). We sort the Hamming distances of samples' hash codes and accordingly gather their latent representations for self-supervised training. Thanks to the recent advances in differentiable sorting approximations, the hash head receives gradients from the sorter so that the hash encoder can be optimized along with the training procedure. Additionally, we describe a novel Sorted Noise-Contrastive Estimation (SortedNCE) loss that selectively picks positive and negative samples for contrastive learning, which allows NSH to mine data semantic relations during training in an unsupervised manner. Our extensive experiments show the proposed NSH model significantly outperforms the existing unsupervised hashing methods on three benchmarked datasets.

\end{abstract}

\section{Introduction}\label{sec_1}


\textit{Learning to hash}~\cite{lsh}, naturally treated as a representation learning task in deep learning, indeed subclasses Approximate Nearest Neighbour (ANN) search that \textit{learns to sort} at scale. The codes' Hamming distances between a query and a bunch of candidates measure their degrees of relevance and further determine their order of presence in the retrieval results. Hence, the conventional evaluation metrics explicitly reflect the positional and order sensitivity of the retrieved candidates in accuracy, which involves an $\operatorname{argsort}$ process, including mean-Average Precision (mAP), top-$k$ accuracy and even Normalized Discounted Cumulative Gain (NDCG) in recommending systems. Oddly enough, most of the existing deep hashing models usually do not implement the concept of sorting, but instead resort to some alternative learning objectives such as recognition and pairwise/triplet losses \cite{cimon}. This counter-intuitive convention raises the question that \textit{can we and shall we train the deep hashing model after sorting to match up with its evaluation metrics?}.

\begin{figure}[t]
	\begin{center}
		\includegraphics[width=\linewidth]{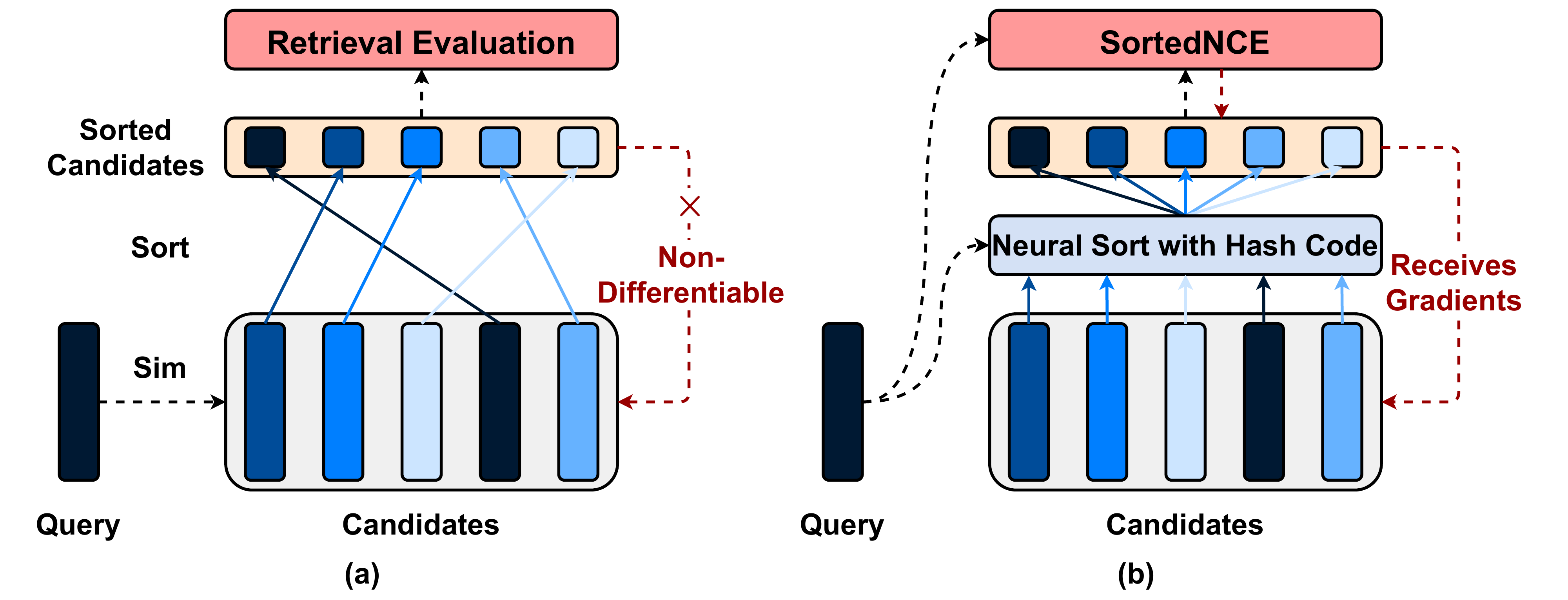}
	\end{center}\vspace{-2ex}
	\caption{A brief motivation of NSH. \textbf{(a)} The actual testing metrics of learning to hash involves non-differentiable $\argsort$ operators. Hence, they can not be directly used for training. \textbf{(b)} The proposed NSH model best mimics the testing procedure that sorts the code similarity with soft approximations and is trained with a list-wise SortedNCE objective end-to-end.}\vspace{-2ex}
	\label{fig_1}
\end{figure}

Recalling one of the best practices in deep learning, the training objective of a model should explicitly represent its ultimate goal or testing measurement. For instance, a segmentation model is usually evaluated by the Intersection over Union (IoU) score and is trained with a similar IoU loss; a density estimation model expects a likelihood objective that describes the sample's density. The coherence between training and testing makes it non-trivial towards better performance. However, in the context of unsupervised hashing, there exist two main challenges to practice this vision.

\paragraph{The Non-Differentiable Kinks in Sorting and ANN} As is discussed above, the main-stream measurements such as mAP and top-$k$ accuracy are based on $\argsort$ operations, which does not derive a differentiable pathway from the actual scores to the hash layers. This makes it impossible to construct an end-to-end deep framework that is trained with the sorted output. We illustrate this problem in \cref{fig_1}~(a). Though the order of relevance can be alternatively converted to positive/negative data pairs that facilitates ranking-based training, this solution, after all, just represents a part of the sorted retrieval list, and it consequently discard some information that potentially matters \cite{ranking}. Additionally, it also leads us to the second question, \ie, how can be determine the order of similarity in an unsupervised manner?

\paragraph{The Lack of Clues of Relevance} Unlike its supervised sibling where similarity labels are off the shelf, unsupervised hashing only observes data themselves. This also hinders existing model from mimicking the sorting-based evaluation metrics during training. Though it is possible to construct pair-wise pseudo labels in an alternating scheme~\cite{sadh,distill}, end-to-end training therefore becomes infeasible for deep models, and the errors of pseudo labels may propagate.

In this paper, we echo the title as the main motivation and tackle the challenges above by proposing Naturally-Sorted Hashing (NSH) that maximally present the properties of sorting during training towards better retrieval performance. As per the first challenge, we adopt the recent advances in differentiable $\softsort$ approximations~\cite{softsort} as one building block that connects the hash encoder on bottom and the sorting-based loss on top. In particular, NSH follows \cite{tbh} that encodes two representations, \ie, a binary hash code for similarity comparison and a continuous latent vector that carries detailed information. The code distances between samples feed the $\softsort$ operator and re-order the latent vectors for self-supervised training. To handle the second challenge, we propose a novel Sorted Noise-Contrastive Estimation (SortedNCE) loss that selectively picks the most related samples as the positive ones for contrastive learning. Hence, NSH mines data semantic relations during training and optimizes sorted retrieval candidate list of each datum on the fly. \cref{fig_1}~(b) briefs our main idea, which better represents the testing scenario of deep hashing. Our main contributions include:
\begin{itemize}
    \item For the first time, we describe an end-to-end NSH model that optimizes the sorted retrieval list that best represents the testing scenario of learning to hash and only requires gradient descent.
    \item To implement this vision, we additionally propose the SortedNCE loss that trains the sorted features in an unsupervised way.
    \item We show the superiority of NSH in retrieval performance on three benchmarked datasets against the most recent unsupervised deep hashing methods. 
\end{itemize}

\section{Related Work}
\paragraph{Unsupervised Hashing}
Early work in unsupervised hashing method mainly focuses on learning compact representations~\cite{itq}. Several recent work with deep learning focus on the hash code quality~\cite{mbe,greedyhash,cimon,hashgan}. Some others pay attention to the semantic awareness of the code \cite{sadh,distill,tbh}, while the majority resort to a pseudo-labelling scheme to mine data similarity as an individual module apart from neural network. Their performance is usually evaluated on the sorted candidates. However, they are not seek to implement the concept of sorting to mine similarity during training.

\paragraph{Hashing with Contrastive Learning}
Contrastive learning is a method to learn the general features of a dataset without labels by construct positive and negative pairs~\cite{hadsell2006dimensionality}. Instance discrimination~\cite{bank} proposes a non-parametric cross-entropy loss to optimize the model at the instance level. Most recent works also uses contrastive learning for hashing ~\cite{cibhash,cimon}. They adopt instance discrimination as the objective where positive and negative instances may still have overlapped semantics. Namely, each query image in a batch only treat its augmented view as a positive sample, which means even for extremely similar samples, they must be pushed apart. In addition, CIBHash \cite{cibhash} and CIMON \cite{cimon} adjust the contrastive loss to suit the hashing learning criterion, yet employing the hash code directly in the contrastive loss does not allow for a good integration of the image's semantic content.

\vspace{-1ex}\section{Preliminaries}\label{sec_2}
\paragraph{Neural Sorting Operators}
Vanilla $\argsort$ is definitely non-differentiable, but recent research finds several approximations that are compatible with neural networks. Though it can be as well viewed as a linear programming problem \cite{fastsort}, we opt to employ a simpler softmax approach \cite{softsort} that determines the permutation of a vector of similarity scores $\s\in\mathbb{R}^N$ as:
\begin{equation}\label{eq_1}
    \softsort(\s) = \softmax \frac{-\ds(\sort(\s)\mathbbm{1}^\intercal, \mathbbm{1}\s^\intercal)}{\tau_\text{s}},
\end{equation}
where $\ds(\cdot)$ is an arbitrary differentiable almost everywhere, semi–metric distance function\footnote{Note that $\ds(\cdot)$ is not the Hamming distance function.}, usually an L-1 norm, and $\tau_\text{s}$ is a temperature hyperparameter.
\paragraph{Twin-Bottleneck Hash Encoder} Though factorized outputs are widely witnessed in deep learning, \cite{tbh} specifies different functionalities two outputs of a hash encoder $\f(\cdot)$ parametrized by $\theta$ that encodes a datum $\xx$
\begin{equation}
    [\hh,\zz]=\f(\xx),~\bb=\operatorname{sign}(\hh).
\end{equation}
$\bb$ is usually followed by a gradient estimator, \eg, $\partial\bb/\partial\hh\coloneqq\mathbb{I}$, to enable end-to-end training. Since the hash code is usually short and less informative, $\hh$ and $\bb$ are only used to compute the pair-wise data similarity that act as the \textit{query}/\textit{key} in the attention mechanism, while $\zz$ carries the detailed information of $\xx$ and plays the role of \textit{value} in attention. Hence, an arbitrary loss built on the top of $\f(\cdot)$ automatically tunes $\bb$ to reflect the semantic locality. We follow the above idea to build the backbone of NSH, but our contributions lie in the operations on the top.

\vspace{-1ex}\section{Method}
NSH considers an unsupervised hashing problem that maps a $d_x$-dimensional datum $\xx\in\{\xx\}_{i=1}^N$ to a binary vector $\bb\in\{-1,1\}^{d_b}$, with $N$ being the size of the entire dataset and $d_b$ being the code length. In this paper, we by default consider an image hashing problem to match up with the conventional experimental settings, but NSH applies to arbitrary data modalities as long as random augmentations apply. 

\begin{figure}[t]
	\begin{center}
		\includegraphics[width=.95\linewidth]{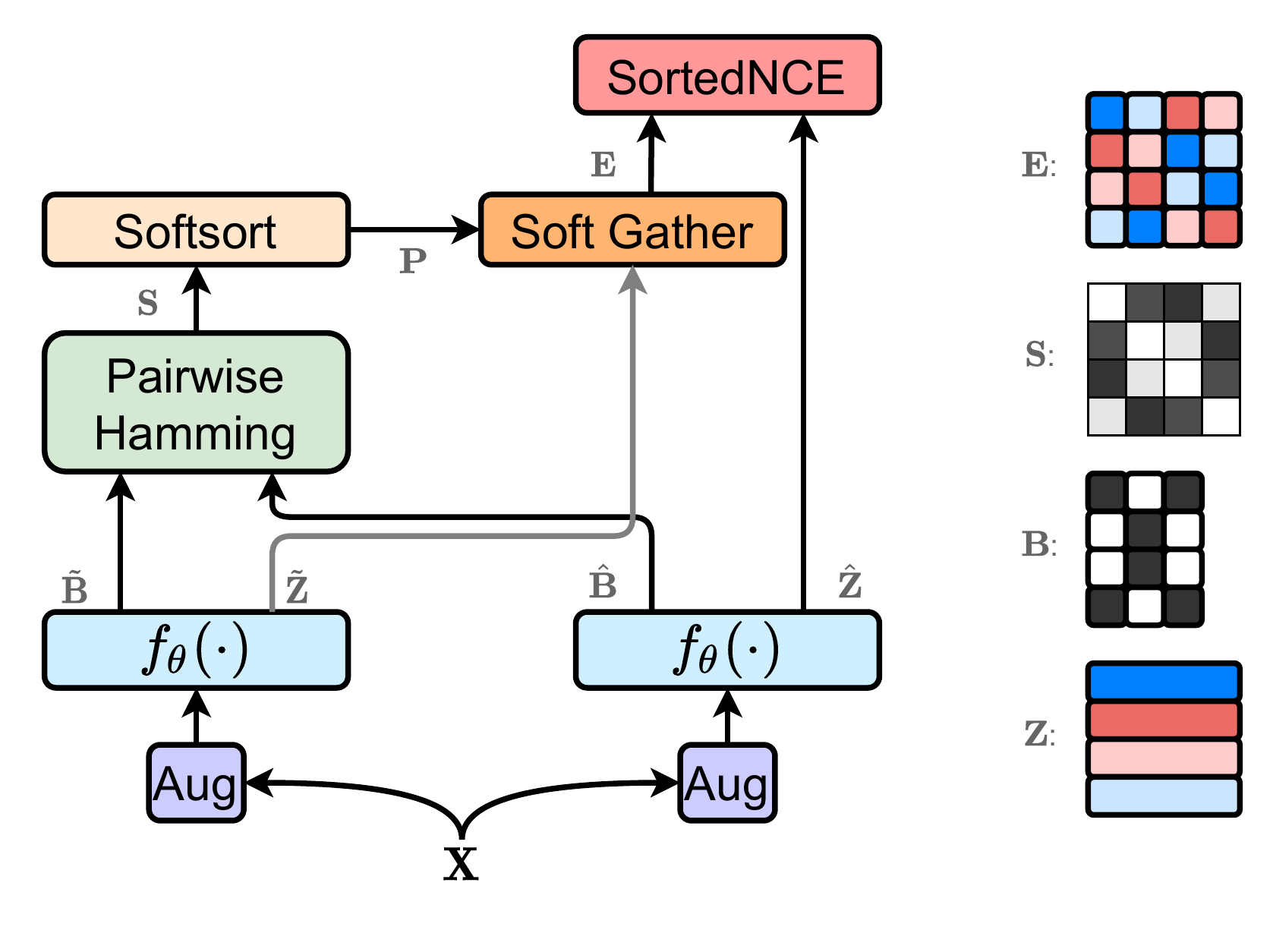}
	\end{center}\vspace{-4ex}
	\caption{The training schematic of NSH. We illustrate the batch-based output shapes on the right where different colors refer to different instances. $\ZZ\in\mathbb{R}^{n\times d_\text{z}}$ are the latent representations. $\BB\in\{-1,1\}^{n\times d_\text{b}}$ are the hash codes. $\SSS\in[0,1]^{n\times n}$ is the code-based similarity matrix for sorting. $\EE\in\mathbb{R}^{n\times n\times d_\text{z}}$ are $n$ soft-sorted lists of representations, of which each is ordered by the Hamming distance to the corresponding instance in the batch for SortedNCE.}\vspace{-1ex}  
	\label{fig_2}
\end{figure}
\vspace{-1ex}\subsection{Model Structure}
As is discussed in Sec.~\ref{sec_2}, we are inspired by \cite{tbh} to employ a twin-bottleneck hashing backbone to produce the actual hash code $\bb$ and another set of latents $\zz$ for each $\xx$ during training, where the loss is imposed on the top of the transformation that both involves $\zz$ and the Hamming distances between $\bb$. NSH also adopts the recent advances in contrastive learning~\cite{simclr} as an unsupervised framework, which requires two sets of independent random data augmentations. In the following, we denote the outputs from the two augmented counterparts using $\tilde{\cdot}$ and $\hat{\cdot}$, \eg, $\tilde{\xx}$ and $\hat{\xx}$, $\tilde{\bb}$ and $\hat{\bb}$, $\tilde{\zz}$ and $\hat{\zz}$, \textit{etc}. In addition, as our training procedure involves multiple instances for sorting, we will use row-ordered batch-wise notations with capital letters when necessary, \eg, $\tilde{\XX}=[\tilde{\xx}_1;\cdots;\tilde{\xx}_n]$, $\hat{\ZZ}=[\hat{\zz}_1;\cdots;\hat{\zz}_n]$, \textit{etc}, with $n$ being the batch size to enable batch-based training.

\paragraph{Training Objective as a Function of Sorting} Our ultimate goal is to build a fully-differentiable model that is trained on a bunch of semantically-sorted candidates:
\begin{equation*}
    \mathcal{L}_\text{Sorted}\coloneqq\cbrace{black}{\chighlight{red!20}{\operatorname{NCE}}}{\color{black}\text{Sec.~\ref{sec_43}}}\color{black}\circ
    \cbrace{black}{\chighlight{cyan!20}{\operatorname{Sort\&Gather}\circ\operatorname{Hamming}}}{\color{black}\text{Sec.~\ref{sec_42}}}\color{black}\circ
    \f(\xx),
\end{equation*}
where each stacked component allows gradient to propagate back to $\f(\cdot)$. \cref{fig_2} gives a glimpse of the structure of NSH. The instances under two different sets of augmentations are both rendered to the backbone encoder $\f(\cdot)$, producing two groups of hash codes ($\tilde{\BB}$ and $\hat{\BB}$) and latents ($\tilde{\ZZ}$ and $\hat{\ZZ}$). The hash codes then determines the way to sort and permute the latents (Sec.~\ref{sec_42}). Then, the permuted latent tensor contributes to the proposed SortedNCE loss for training (Sec.~\ref{sec_43}), while the training procedure is given in Sec.~\ref{sec_44}.

\vspace{-1ex}\subsection{Sorting and Gathering the Latents}\label{sec_42}
Similar to \cite{tbh}, we implant the computation of pair-wise hash code distances in the training model so that the code similarity can be optimized by any losses on the top. Formally, we define the code-based affinity matrix $\SSS$ as:
\begin{equation}\label{eq_3}
    \SSS=\BBa\BBb^\intercal/2d_b +0.5,
\end{equation}
so that its each entry represents the normalized pair-wise similarity, \ie, $\SSS[i,j]=1-\text{Hamming}(\tilde{\bb}_i, \hat{\bb}_j)/d_b$. \cref{eq_3} only involves linear operations that are differentiable everywhere so that it can be used as a building block in neural network for back-propagation. Note that we are \textbf{not} imposing any loss term on $\SSS$ to prevent the model from falling into the pseudo-labelling training scheme, as discussed in Sec.~\ref{sec_1}

\paragraph{Sort out the Relevance with Hash Codes}
As the key operation of NSH, we treat each sample in the batch as the retrieval query, while letting the whole batch as the candidates. For each datum, one can pick the most relevant ones and place it at the head of the retrieval sequence and so on. To reflect this procedure in the model, we recall \cref{eq_1} to compute the sort-permutation matrix $\pp_i$ for each $\xx_i$, \ie,
\begin{equation}\label{eq_4}
    \pp_i = \softsort(-\sss_i)\in (0,1) ^{n\times n},
\end{equation}
where $\sss_i\in [0,1]^n$ is the $i$-th row of $\SSS$ (\cref{eq_3}) that describes the relevance of the $i$-th item in the batch to all the others. Namely, an entry of $\pp_{i}[j,k]$ at the position of $[j,k]$ stands for the probability of $\xx_k$ becoming the most $j$-th related item to $\xx_i$~\cite{softsort}.

\paragraph{Re-order Features by Soft Gathering/Permutation} For each $\xx_i$, NSH produces a matrix of candidates' embeddings in the batch $\ee_i$ where the features of the most similar items are placed on the top. As now we already have the soft-permutation matrix from \cref{eq_4}, $\ee_i$ can be defined by the following soft-gathering process:
\begin{equation}\label{eq_5}
    \ee_i=\pp_i\ZZa\in\mathbb{R}^{n\times d_z}.
\end{equation}
Here $d_z$ is the size of the continuous latents $\tilde{\zz}$. To clarify, $\ee_i$ is just the sorted embedding matrix for $\xx_i$, while a batch of data makes it a three-dimensional tensor, \ie, $\EE =[\ee_1;\cdots;\ee_n]\in\mathbb{R}^{n\times n\times d_z}$. The batch-based implementation of \cref{eq_5} can be easily achieved by the Einstein summation convention in recent deep learning toolboxes (see Alg.~\ref{alg}).

\paragraph{Remark 1: The Order of $\EE$ Matters.}
$\EE$ is literally $n$ sorted lists of the latents $\ZZa$, of which each list $\ee_i$ contains the representations of the in-batch retrieval results of $\xx_i$ in a descending order of similarity. Notably, as $\tilde{\bb}_i$ and $\hat{\bb}_i$ are expected to be identical, the first row of $\ee_i$, \ie, $\ee_i[1,:]$, will naturally be close to $\tilde{\zz}_i$, representing $\xx_i$ itself as the most relevant one in the batch. From the second row on, the relevance decreases. The steps defined by \cref{eq_3,eq_4,eq_5} automatically selects semantically-related instance and place them at the beginning of $\EE$. Hence, one can easily determine the positive/negative samples of each $\xx_i$ in a fully-differentiable way. This procedure mimics the real testing scenario of retrieval and enables list-wise training for better performance.

\paragraph{Remark 2: Why Do We Gather $\ZZ$?} It is possible that we construct a single-bottleneck model and replacing $\ZZa$ by $\BBa$ in \cref{eq_5}. However, hash codes are usually short and less informative. Though they are able to encode coarse-granular semantic similarity, it would be hard for them to carry the full identity of each datum against the others. We experimentally show that this leads to sub-optimal performance when trained with contrastive learning objectives that heavily rely on identity preservation. We follow \cite{tbh} to use another continuous representation here to favour the learning objective. Alternatively, our hash codes can be viewed as the attention head that permutes $\ZZ$, the loss built on the top automatically promotes the semantic awareness of the codes.

\vspace{-1ex}\subsection{SortedNCE}\label{sec_43}
Many existing unsupervised hashing models struggle in determining the similarity between data points, and they usually needs a held-out pseudo-labelling step that introduce additional noise~\cite{greedyhash,cimon}. However, NSH already has a bunch of sorted features $\EE$ off the shelf, which makes it extremely easy to determine positive samples.

On the other hand, there would be multiple samples that share the same semantic information, but the vanilla single-label InfoNCE~\cite{infonce} loss does not consider this overlap of semantics. We propose a learning objective that works with multiple positive pairs. In particular, since $\EE$ is ordered, we define the SortedNCE loss based on the positions of the logits and let the first $m$ samples being positive, \ie,
\begin{equation}\label{eq_6}
\begin{split}
   \mathcal{L}_{\text{Sorted}}&=\frac{-1}{mn}\sum_{i=1}^n\sum_{j=1}^m\\
    &\color{black}\underbrace{\color{black}\log\frac{\kappa(\ee_i[\highlight{j},:],\hat{\zz}_i)}
    {\kappa(\ee_i[\highlight{j},:],\hat{\zz}_i)+\sum_{k=m+1}^n\kappa(\ee_i[\highlight{k},:],\hat{\zz}_i)}}_{\color{BrickRed}\text{Positive and negative samples only depend on the sorting positions}},
\end{split}
\end{equation}
where $\kappa(\mathbf{a},\mathbf{b})=\exp(\cos(\mathbf{a}, \mathbf{b})/\tau_\text{c})$. $m$ and $\tau_\text{c}$ are treated as hyperparameters. Intuitively, \cref{eq_6} constructs $m$ contrastive for $m$ positive candidates in $\ee_i$, of which each shapes a single-label cross-entropy term. Importantly, since the features of positive samples have fixed positions of $1\cdots m$ in $\ee_i$, NSH does not requires an additional $\argsort$ operator to find the most relevant items that involves non-differentiable computations. We discuss the benefits of this design in Sec.~\ref{sec_45}. A single $m$-label cross-entropy term may work here too, by treating the first $m$ logits as positive. This solution is similar to SupCon \cite{supcon}, but we experimentally show in Sec. \ref{sec_abl} that it underperforms SortedNCE for our case.

Also note that \cref{eq_6} is not a universal loss, and it is dedicated to retrieval models such as NSH that contains the sort sorting and gathering layers described by \cref{eq_3,eq_4,eq_5}. The time complexity of SortedNCE is $O(mn)$ for each sample. Considering the fact that $m$ is usually small, this loss would not significantly increase the time of training.

\vspace{-1ex}\subsection{Training and Inference}\label{sec_44}

Training NSH is extremely simple. In addition to $\mathcal{L}_\text{Sorted}$, it only requires the very conventional quantization loss $\mathcal{L}_\text{R}=(\|\operatorname{sg}(\BBa)-\HHa\|_2+\|\operatorname{sg}(\BBb)-\HHb\|_2)/2n$ \cite{dh} to enhance the concreteness of the hashing layer, \ie,
\begin{equation}
    \mathcal{L}_\text{NSH}=\mathcal{L}_\text{Sorted}+\mathcal{L}_\text{R}.
\end{equation}
Here, $\operatorname{sg}(\cdot)$ is the stop-gradient operation. We place $\operatorname{sg}(\cdot)$ here to avoid duplicated gradients from $\BB$ since we have previously manually defined the gradient estimator of $\partial\bb/\partial\hh$. Alg.~\ref{alg} describes the training process of NSH in Python-style pseudo codes. 

\paragraph{Encoding Testing Samples}
\setlength{\textfloatsep}{0pt}
\begin{algorithm}[t]
	\small
	\caption{The Training Procedure of NSH}
	\label{alg}
	\textbf{Input:}\hspace{0mm} Dataset $\mathcal{D}=\{\xx_i\}_{i=1}^N$ and batch size $n$.\\
	\textbf{Output:}\hspace{0mm} Network parameters $\theta$.\\

	\For{$\mathtt{batch~in}~\mathcal{D}\func{.repeat}()$}{
	    $\variable{batch1}=\func{aug}(\variable{batch})~\comment{[n~d_x]}$\\
	    $\variable{batch2}=\func{aug}(\variable{batch})$\\
	    $\variable{[b1,~z1],~[b2,~z2]} = \func{f_\theta}(\variable{batch1}),~\func{f_\theta}(\variable{batch2})$\\
	    $\variable{s}=\func{matmul}(\variable{b1,b2.T})/2/\variable{d_l} + 0.5~\comment{1-Hamming/{d_l}}$\\
	    $\variable{p}=\func{softsort}(\variable{-s})~\comment{[n~n~n]}$\\
	    $\variable{e}=\func{einsum}(\param{'nnn,nd\rightarrow nnd'},~\variable{p,~z1})$\\
	    $\comment{The~code~below~describes~SortedNCE}$\\
	    $\variable{labels}=\func{onehot}(\func{zeros}(\variable{[n]}),~\variable{n{-}m{+}1})$\\
	    $\variable{cos}=\func{einsum}(\param{'nnd,nd\rightarrow nn'},~\variable{e,~z2})$\\
	    $\variable{loss}=0$\\
	    \For{$\mathtt{i~in}~\func{range}(\variable{m})$}{
	    
	    $\variable{pos,~neg}=\variable{cos[:, i],~cos[:, {m}{:}]}$\\
	    $\variable{logits}=\func{softmax}(\func{concat}(\variable{[pos,~neg]})/\variable{\tau_c})$\\
	    $\variable{loss}~{+}{=}~\func{cross\_entropy}(\variable{logits,~labels})/\variable{m}/\variable{n}$
	    }
	    $\variable{loss}~{+}{=}~\func{quantization\_loss}$\\
	    $\func{optimizer.apply\_gradients}(\variable{loss,~\theta})$
	}
\end{algorithm}
On testing, NSH does not require any sorting or gathering layers. The hash code of a testing sample can be directly obtained from the backbone $\f(\cdot)$ without interacting with the other ones, while the semantic awareness is memorized by the parameters during training.

\vspace{-1ex}\subsection{Discussion: What Makes It Different?}\label{sec_45}
\begin{table*}[t]
	\begin{center}
		\small
		\resizebox{\textwidth}{!}{
			\begin{tabular}{l l ccc  ccc  ccc}
				\hline
				\multirow{2}{*}{\textbf{Method}}&\multirow{2}{*}{\textbf{Reference}}&\multicolumn{3}{c}{\textbf{ CIFAR-10}}&\multicolumn{3}{c}{\textbf{NUS-WIDE} }&\multicolumn{3}{c}{\textbf{MS COCO}}\\\cline{3-11}
				&& 16 bits& 32 bits & 64 bits& 16 bits& 32 bits & 64 bits& 16 bits& 32 bits & 64 bits\\ \hline\hline
				AGH~\cite{agh}&ICML11& 0.333& 0.357 & 0.358 & 0.592& 0.615& 0.616&0.596 &0.625 & 0.631\\
				ITQ~\cite{itq}&PAMI13& 0.305& 0.325& 0.349& 0.627& 0.645& 0.664& 0.598& 0.624& 0.648\\
				DGH~\cite{dgh}&NeurIPS14& 0.335& 0.353& 0.361& 0.572& 0.607& 0.627& 0.613& 0.631&0.638\\\hline
				DeepBit~\cite{deepbit}&CVPR16& 0.194& 0.249& 0.277& 0.392& 0.403& 0.429& 0.407& 0.419 & 0.430 \\
				SGH~\cite{sgh}&ICML17 & 0.435& 0.437 & 0.433& 0.593& 0.590& 0.607& 0.594& 0.610& 0.618\\
				BGAN~\cite{bgan}&AAAI18& 0.525& 0.531& 0.562& 0.684& 0.714& 0.730& 0.645& 0.682& 0.707\\
				BinGAN~\cite{bingan}&NeurIPS18& 0.476& 0.512& 0.520& 0.654& 0.709& 0.713& 0.651& 0.673& 0.696\\
				GreedyHash~\cite{greedyhash}&NeurIPS18& 0.448& 0.473& 0.501& 0.633& 0.691& 0.731 & 0.582& 0.668& 0.710\\
				HashGAN~\cite{hashgan}&CVPR18& 0.447& 0.463& 0.481& -& -& -& -& -& -\\
				DVB~\cite{dvbj}&IJCV19& 0.403& 0.422& 0.446& 0.604& 0.632& 0.665& 0.570& 0.629& 0.623\\
				DistillHash~\cite{distill}& CVPR19& 0.284& 0.285& 0.288& 0.667& 0.675& 0.677& -& -& -\\
				{TBH}~\cite{tbh}& CVPR20&{0.532}&{0.573}&{0.578}&{0.717}&{0.725}&{0.735}&{0.706}&{0.735}&{0.722}\\
				{$\rm MLS^{3}RDUH$}~\cite{tu2020mls3rduh}& IJCAI20&{0.369}&{0.394}&{0.412}&{0.713}&{0.727}&{0.750}&{0.607}&{0.622}&{0.641}\\
				{DATE}~\cite{date}& MM21&{0.577}&{0.629}&{0.647}&\textbf{0.793}&{0.809}&{0.815}& -& -&-\\
				{MBE}~\cite{mbe}& AAAI21&{0.561}&{0.576}&{0.595}&{0.651}&{0.663}&{0.673}& -& -&-\\
				{CIMON}~\cite{cimon}*& IJCAI21&{0.451}&{0.472}&{0.494}& -& -&-&-&-&-\\
				{CIBHash}~\cite{cibhash}& IJCAI21&{0.590}&{0.622}&{0.641}&{0.790}&{0.807}&{0.815}&{0.737}&{0.760}&{0.775}\\\hline
				\textbf{NSH}& \textbf{Proposed}&\textbf{0.706}& \textbf{0.733}& \textbf{0.756}& {0.758}& \textbf{0.811}&\textbf{0.824}&\textbf{0.746}&\textbf{0.774}&\textbf{0.783}
				\\\hline
			\end{tabular}
		}\vspace{-2ex}
		\caption{Performance comparison (\wrt~mAP) of NSH and the state-of-the-art \textbf{unsupervised} hashing methods. *Note that we use a more common setting on NUS-WIDE with the 21 most frequent classes, while some papers report results on 10 classes.
	}\label{tab_map}
	\end{center}\vspace{-3ex}
\end{table*}
NSH defines a specific gradient pathway from the contrastive loss to the hash codes that reflect the rewards to both the code semantic similarity and the sorted candidates' quality end-to-end. We slightly abuse the differential notations to represent the chain rules of these concepts as follows:
\begin{equation}\label{eq_8}
\frac{\partial \mathcal{L}_\text{Sorted}}{\partial \BB}=\cbrace{white}{\chighlight{red!20}{\frac{\partial \mathcal{L}_\text{Sorted}}{\partial \PP}}}{\color{BrickRed}\substack{\text{Sorting/Retrieval}\\\text{Rewards}}}
\cbrace{white}{\chighlight{cyan!20}{\frac{\partial \PP}{\partial \SSS}}}{\color{RoyalBlue}\substack{\text{Similarity}\\\text{Rewards}}}
\cbrace{white}{\chighlight{gray!20}{\frac{\partial \SSS}{\partial \BB}}}{\color{black!80}\substack{\text{Code}\\\text{Quality}}}
\color{black}.
\end{equation}
In this way, the model automatically optimizes the key components of hashing task to best fit and favour the presence of $\mathcal{L}_\text{Sorted}$, even if the similarity ground truth is not given.

\paragraph{As to a Hard-Sort Baseline} One can easily build a baseline by replacing \cref{eq_4,eq_5} with $\argsort$ and computing the contrastive loss according to the highest $\argsort$ entries. However, we argue this design is just enhancing the decisions of $\argsort$, because the absence of the gradients $\partial\argsort/\partial\SSS$ does not allow the model to mine the data similarity and fail to update $\SSS$ during gradient descent.

\paragraph{As to Many Existing Models} Some methods does not consider optimizing $\SSS$ during training~\cite{greedyhash,cimon}, which may lead to biased results. The majority of recent works treat $\SSS$ as pseudo labels that are updated out of the training loop, since their design do not compute the gradients of $\partial\mathcal{L}/\partial\SSS$ of \cref{eq_8}~\cite{sadh,distill}. As discussed previously, this design allows the errors caused by false assignments to propagate during network training. Last but not least, NSH is the first model to optimize the mocked sorting results through $\partial\mathcal{L}/\partial\PP$ of \cref{eq_8}, which literally performs unsupervised list-wise training end-to-end. Echoing our motivation, this strategy better fits the retrieval task that undergoes list-wise evaluation measurements.

\vspace{-2ex}\section{Experiments}
\vspace{-1ex}\subsection{Experimental Setup}

\noindent\textbf{CIFAR-10~\cite{cifar}} comes with 60,000 images. We follow \cite{hashgan} to have a 50,000-10,000 train-test split.

\noindent\textbf{NUS-WIDE~\cite{nus}} has of 81 categories of images. We adopt the 21-class subset following \cite{cibhash}. 100 images of each class are utilized as a query set, with the remaining being the gallery.

\noindent\textbf{MS COCO~\cite{coco}} is a benchmark for multiple tasks. We use the conventional set with 12,2218 images. We randomly select 5,000 images as queries with the remaining ones the database.

\noindent\textbf{Evaluation Metric}~~
We adopt several widely-used evaluation metrics, including mean Average Precision (mAP), top-$k$ precision (P@$k$), Precision-Recall (P-R) curves and precision of Hamming radius within 2 (P@r=2). Following the recent convention~\cite{tbh,cibhash}, we adopt mAP@1000 for CIFAR-10, mAP@5000 for NUS-WIDE and MSCOCO. For all datasets, two data points will be considered as relevant if they share at least one common label. 

\begin{table}[t]
	
	\small
	\resizebox{\linewidth}{!}{
		\begin{tabular}{l ccc ccc}
			\hline
			\multirow{2}{*}{\textbf{Method}} &\multicolumn{3}{c}{\textbf{ CIFAR-10}}&\multicolumn{3}{c}{\textbf{MS COCO}}\\\cline{2-7}
			&16 bits& 32 bits & 64 bits& 16 bits& 32 bits & 64 bits\\\hline\hline
			ITQ & 0.276& 0.292 & 0.309 & 0.607& 0.637& 0.662\\
			AGH & 0.306& 0.321& 0.317& 0.602& 0.635& 0.644\\
			DGH & 0.315& 0.323& 0.324 & 0.623& 0.642& 0.650\\
			HashGAN & 0.418& 0.436& 0.455 &-&-&-\\
			SGH & 0.387 & 0.380 & 0.367 & 0.604 & 0.615 & 0.637\\
			GreedyHash & 0.322 & 0.403 & 0.444 & 0.603 & 0.624 & 0.675 \\
			{TBH} & {0.497}& {0.524}& {0.529}& {0.646}& {0.698}& {0.701}\\
			CIBHash& 0.526& 0.570& 0.583& 0.734& 0.767& 0.785\\
			\hline
			\textbf{NSH}&\textbf{0.691}&\textbf{0.716}&\textbf{0.744}& \textbf{0.733}& \textbf{0.770}& \textbf{0.805}\\
			\hline
		\end{tabular}
	}\vspace{-1ex}
\caption{P@1000 results of NSH and compared methods on CIFAR-10 and MS COCO.}\vspace{-2ex}
\label{tab_pre}\vspace{3ex}
\end{table}

\noindent\textbf{Implementation Details}~~The proposed method is implemented with Tensorflow. We use the Adam optimizer \cite{adam} to train the networks with a learning rate of $1\times10^{-5}$, and the batch size is 50. We train the model for 200 epochs at most. All the images are resized to $224\times224\times3$ and we adopt the image augmentation strategies of MoCo-v2~\cite{mocov2}. We use the ResNet-50~\cite{resnet} until the last pooling layer and top two fully-connected layers as the hash head and the latent feature head. The hash head is followed by $\operatorname{tanh}$ and $\operatorname{sign}$ operations to produce $\bb$, while the latent feature head is followed by a L-2 normalization layer to produce $\zz$, with a dimensionality of $d_\text{z}=1024$. The contrastive temperature $\tau_\text{c}$ and the number of positive samples $m$ we picked was set to \{0.1, 0.5, 0.5\} and \{2, 3, 3\} for CIFAR-10, NUS-WIDE and MS COCO. Following \cite{softsort}, the $\softsort$ temperature is set to the code length $\tau_\text{s}=d_\text{b}$.

\vspace{-1ex}\subsection{Comparison with the SotA}
\noindent\textbf{Baselines}~ We compare NSH against 17 state-of-the-art baselines, including 3 traditional unsupervised hashing methods and 14 recent unsupervised hashing methods. For a fair comparison, we report the results with  VGG features~\cite{vgg} which is pretrained on ImageNet if the baseline is not trained from scratch.

\begin{figure}[t]
	\begin{center}
		\includegraphics[width=\linewidth]{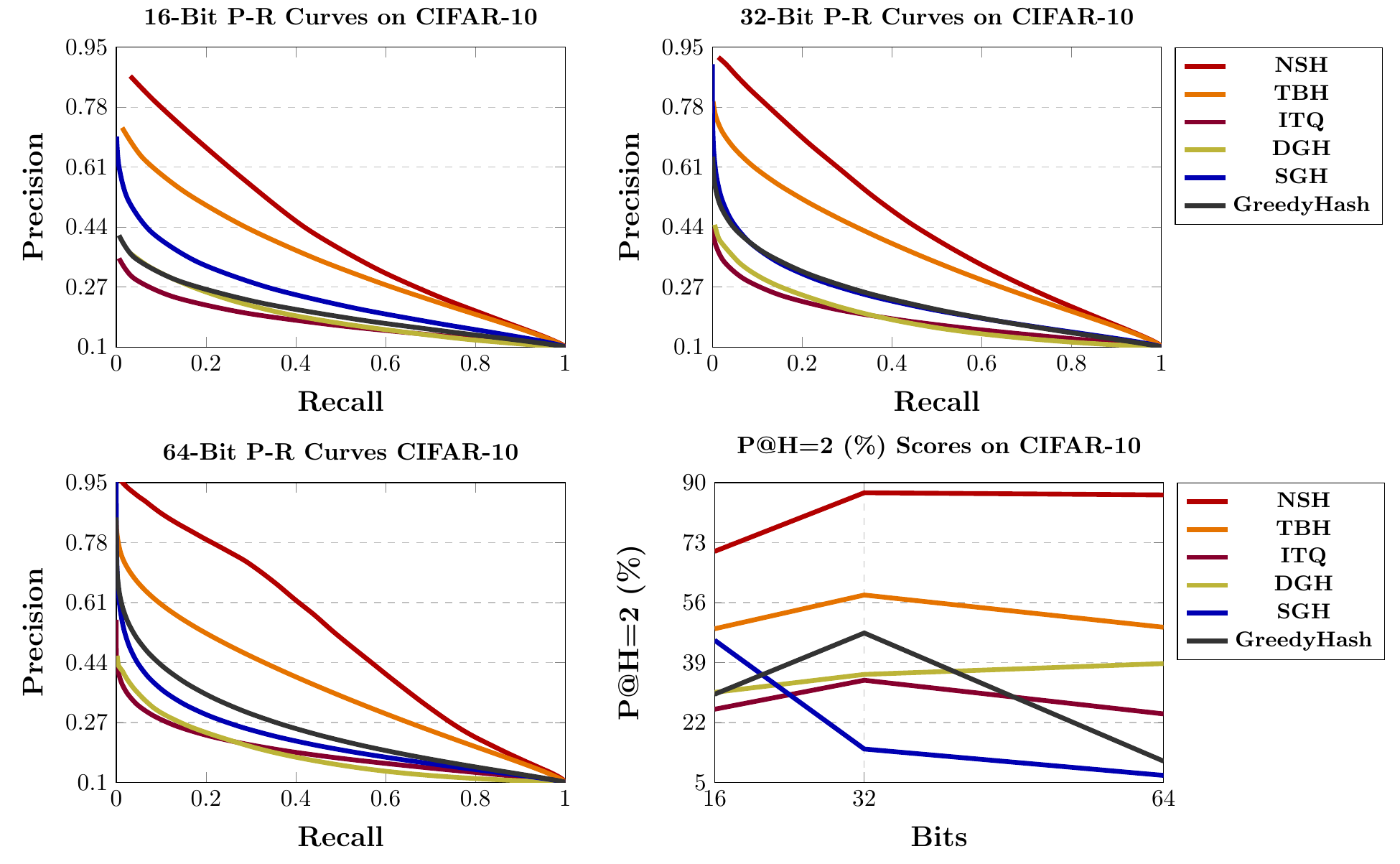}        
	\end{center}\vspace{-2ex}
	\caption{P-R curves and P@H$\leq$2 results of NSH and compared methods on CIFAR-10.}\vspace{-1ex}
	\label{fig_pr}
\end{figure}

\noindent\textbf{Results}~ Tab.~\ref{tab_map} shows the retrieval performance in mAP. It can be clearly observed that NSH obtain the best results on the three datasets. Another interesting observation is that NSH is significantly better than CIBHash across different hash bits and datasets. Note that both these two methods employ contrastive learning. 
In addition, the P-R curves and the precision within Hamming radius of 2 (P@H=2) of NSH and several baselines on CIFAR-10 are reported in \cref{fig_pr}.
\begin{table}
	
	\small
	\resizebox{0.99\linewidth}{!}{
		\begin{tabular}{ll ccc}
			\hline
			&\textbf{Baseline}& \textbf{16 bits}& \textbf{32 bits}& \textbf{64 bits}\\\hline\hline
			\ref{bl_1}&Hard-Sort Baseline & 0.323& 0.405& 0.477\\
			\ref{bl_2}&Without $\mathcal{L}_{R}$& 0.686 & 0.718 & 0.735\\
			\ref{bl_3}&$\mathcal{L}_\text{Sorted}\rightarrow$ Decoding Loss & 0.489& 0.501& 0.544\\
			\ref{bl_4}&Single Bottleneck& 0.606& 0.631& 0.650\\
			\ref{bl_5}&Without $\softsort$ &  0.641& 0.695& 0.707\\
			\ref{bl_6}&$\mathcal{L}_\text{Sorted}\rightarrow$Multi-Label NCE& 0.689& 0.710& 0.730 \\\hline
			&\textbf{NSH}& \textbf{0.706}& \textbf{0.733}& \textbf{0.756}\\\hline
		\end{tabular}
	}\vspace{-2ex}
\caption{Ablation study results of mAP@1000 on CIFAR-10. The baselines are constructed by replacing some key modules of NSH.}\label{tab_abl}\vspace{1ex}
\end{table}
\vspace{-1ex}\subsection{Ablation Study}\label{sec_abl}
We validated the effectiveness of our motivation and design via the following baselines, with the results shown in Tab.~\ref{tab_abl}.
\vspace{-0ex}\begin{enumerate}[label=\color{red!70!black}(\roman*),wide,labelindent=0pt]
\item\textbf{Hard-Sort Baseline.}\label{bl_1} We first explore the effect of our core motivation with differentiable sort on hashing. This baseline is also described in Sec.~\ref{sec_45} that replaces \cref{eq_4,eq_5} with $\argsort$ and computes the contrastive loss according to the highest $\argsort$ entries. As is previously discussed, this baseline fail to optimize the code during training, and thus the results are not promising.

\vspace{-0.5ex}\item\textbf{Without $\mathcal{L}_{R}$.}\label{bl_2} We also evaluate how important the convention quantization loss can be in NSH. From Tab.~\ref{tab_abl}, we see that this traditional regularizer dose not influence the final results much. Hence we can give full credit to our design for the good performance.

\vspace{-0.5ex}\item\textbf{$\mathcal{L}_\text{Sorted}\rightarrow$Decoding Loss.}\label{bl_3} To demonstrate the effectiveness of our SortedNCE, we construct this baseline by removing $\mathcal{L}_\text{Sorted}$ and add a decoder with a decoding loss after sorting. In this case, reconstruction only requires $\ee_i[1,:]$ as the most relevant latents, but the network is still fully trainable. It can be observed that this baseline obtains similar performance to TBH \cite{tbh} as both of them involve an auto-encoding structure.

\vspace{-0.5ex}\item\textbf{Single Bottleneck.}\label{bl_4} This baseline removes $\zz$ in the network, so that \cref{eq_5} gathers the hash code only, \ie, $\ee_i=\pp_i\BBa$. This baseline performs close to CIBHash \cite{cibhash} as both of their contrastive losses are imposed to the code-based features. The drop in performance accords our intuition to employ the twin-bottleneck encoder.

\vspace{-0.5ex}\item\textbf{Without $\softsort$.}\label{bl_5} This baseline removes the operations defined by \cref{eq_4,eq_5}, and then compute $\tilde{\EE}=\SSS\BBa, \hat{\EE}=\SSS\BBb$. Hence, a conventional SimCLR-like contrastive learning loss \cite{simclr} can be built upon $\tilde{\EE}$ and $\hat{\EE}$. Though it produces good results as well, the performance margin between this baseline and NSH is still significant, showing that our core motivation to train a sorted list is valid.

\vspace{-0.5ex}\item\textbf{$\mathcal{L}_\text{Sorted}\rightarrow$Multi-Label NCE.}\label{bl_6} Our SortedNCE actually constructs $m$ cross-entropy terms. Each one only has one positive label. It is also possible to replace it with a multi-label NCE loss such as SupCon \cite{supcon}, by marking the first $m$ entries as positive. However, this baseline still underperforms NSH. We suspect this is because the $\softmax$ operator fits single-label objective better.

\end{enumerate}
\begin{figure}[t]
	\begin{center}
		\includegraphics[width=\linewidth]{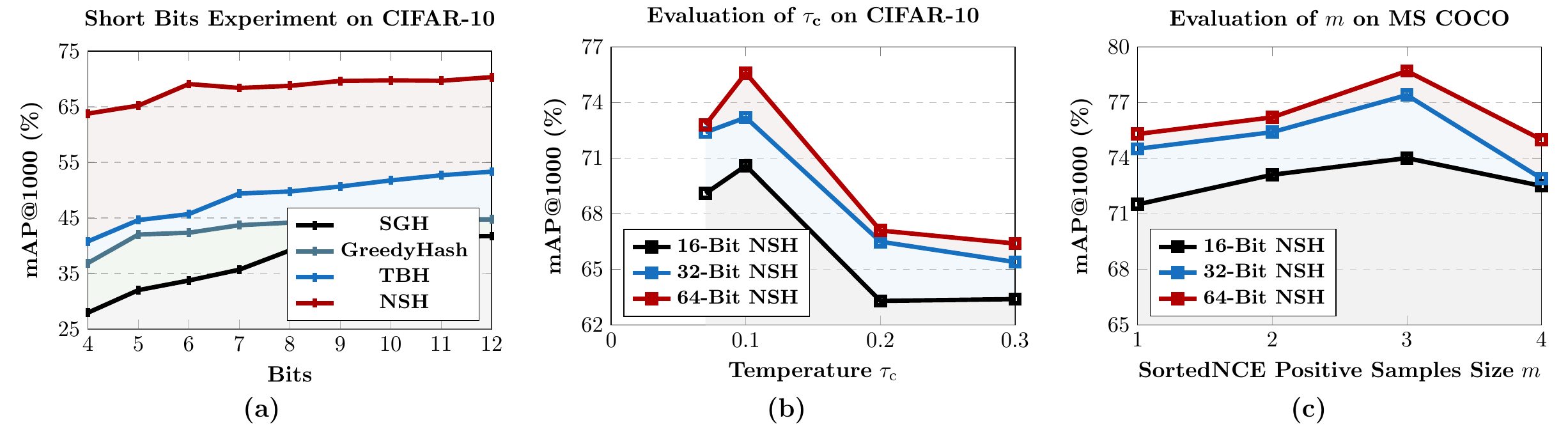}
	\end{center}
	\vspace{-3ex}
	\caption{\textbf{(a)} mAP@1000 results with extremely short code lengths on CIFAR-10. \textbf{(b)} Effects of different temperatures $\tau_\text{c}$. \textbf{(c)} Effects of different sizes of $m$ in SortedNCE.}\vspace{-2ex}
	\label{fig_hp}\vspace{0ex}
\end{figure}
\vspace{-1ex}\subsection{Hyperparameters and Visualization}
\begin{figure}[t]
	\begin{center}
		\includegraphics[width=\linewidth]{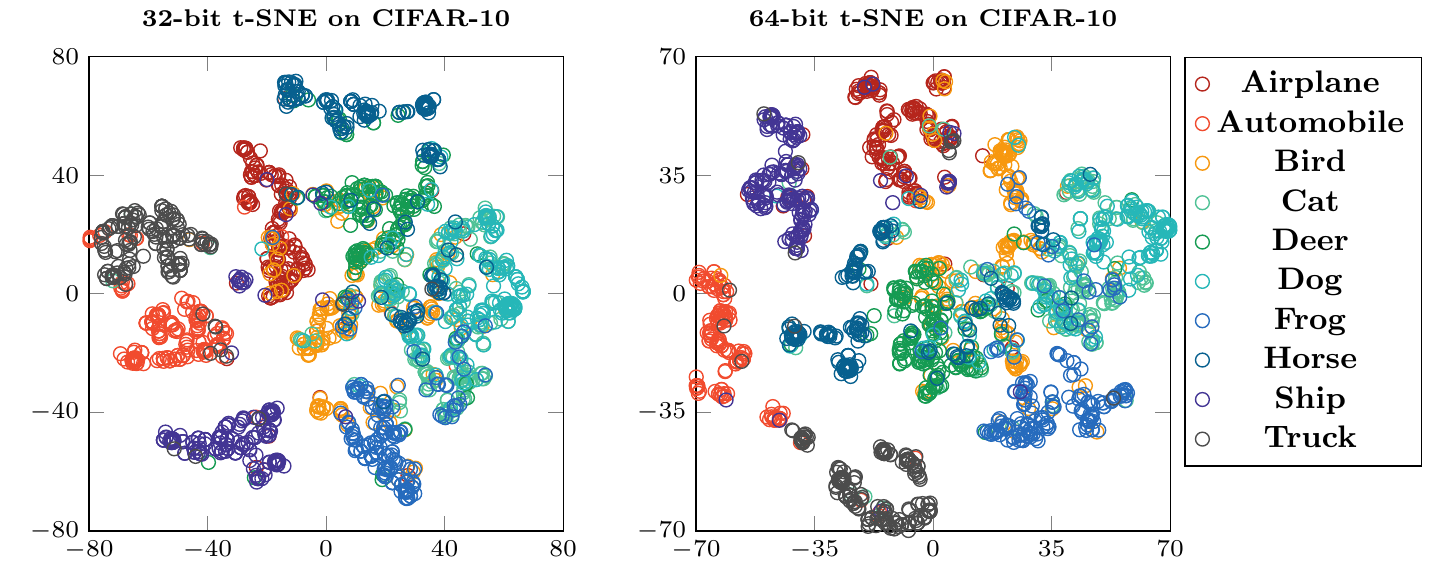}
	\end{center}
	\vspace{-3ex}
	\caption{32-bit and 64-bit t-SNE visualization results on CIFAR-10}\vspace{2ex}
	\label{fig_tsne}
\end{figure}

We study the influence of the temperature $\tau_\text{c}$ and the number of positive samples $m$ we picked, which are reported in \cref{fig_hp} (b) and (c). We also provide the results with extremely short code length in \cref{fig_hp} (a). Though the performance under different settings of hyperparameters varies, it is overall stable and is yet representing the state-of-the-art. NSH is not very sensitive to the hyperparameters. We do not assess different values of $\tau_\text{s}$ as its value is recommended by \cite{softsort}. In addition, the proportion of $\mathcal{L}_\text{R}$ does not influence the performance much so we skip its weighing hyperparameter here. We plot the t-SNE \cite{tsne} results in \cref{fig_tsne} to illustrate our semantic awareness.

.

\vspace{-2ex}\section{Conclusion}\vspace{-0.5ex}
In this paper, we discussed the disagreement in the presence of training and testing objectives of unsupervised deep hashing and then proposed NSH to solve this problem. NSH overcame the main difficulties to mine the data semantics by sorting and is trained in a list-wise fully-differentiable manner that better reflects the testing scenario of retrieval. We adopted the recent advances in $\softsort$ and proposed SortedNCE to implement our vision. Our experimental results endorsed our motivation and design, showing the superiority of NSH in performance.

{\small
\bibliographystyle{named}
\bibliography{refs}}
\end{document}